\pdfoutput=1

\documentclass[11pt]{article}

\usepackage{emnlp2021}

\usepackage{times}
\usepackage{latexsym}

\usepackage[T1]{fontenc}

\usepackage[utf8]{inputenc}

\usepackage{microtype}

\usepackage{amsmath}
\usepackage{amsfonts}
\usepackage{bm}
\usepackage{comment}
\usepackage{booktabs}
\usepackage{multirow}
\usepackage{multicol}
\usepackage{tabularx}

\newcommand{\kevin}[1]{}
\newcommand{\davis}[1]{}

\newcommand{\hstate}[2]{\bm{h}_{#1:#2}}
\newcommand{\modstate}[2]{\tilde{\bm{h}}_{#1:#2}}

\title{Dynamic memory evaluation}
\title{Optimizing Transformer hidden states for improved language model adaptation}
\title{Adapting hidden states for fast (/low data) language model adaptation}
\title{Adapt to your prompt by optimizing (language model)? hidden states}
\title{Reconsider/Rethink/Recompute/ your prompt/context (with dynamic hidden states|by optimizing (language model)? hidden states)}
\title{Rapid Prompt Adaptation by Optimizing Language Model Hidden States}
\title{Context Reconsideration: Optimizing Hidden States for Rapid Adaptation of Language Models}
\title{Optimizing Hidden States for Language Model Adaptation}
\title{ReCon: Reconsidering the Context by Optimizing Language Model Hidden States}
\title{Prompt-Driven Optimization of Language Model Hidden States}
\title{Prompt-Driven Optimization of Hidden States for Improved Language Modeling}
\title{Prompt Adaptation by Optimizing Language Model Hidden `States}
\title{Reconsider the Context: Optimizing Hidden States for Language Models}
\title{Reconsider the Context: Optimizing Language Model Hidden States}
\title{Reconsidering the Past: Optimizing Language Model Hidden States for Observations}
\title{Reconsidering the Past: Observation-Driven Optimization of Language Model Hidden States}
\title{Reconsidering the Past: Optimizing Hidden States in Language Models}

\author{Davis Yoshida \\
  Affiliation / Address line 1 \\
  Affiliation / Address line 2 \\
  Affiliation / Address line 3 \\
  \texttt{dyoshida@ttic.edu} \\\And
  Kevin Gimpel \\
  Affiliation / Address line 1 \\
  Affiliation / Address line 2 \\
  Affiliation / Address line 3 \\
  \texttt{kgimpel@ttic.edu} \\}

\author{Davis Yoshida \ \quad\quad\quad\  Kevin Gimpel \\
     Toyota Technological Institute at Chicago, IL, USA, 60637 \\ \{\texttt{dyoshida,kgimpel}\}\texttt{@ttic.edu}}

\begin{document}
\maketitle

\begin{abstract}
We present Hidden-State Optimization (HSO), a gradient-based method for improving the performance of transformer language models at inference time. Similar to dynamic evaluation \citep{krause2018dynamic}, HSO computes the gradient of the log-probability the language model assigns to an evaluation text, but uses it to update the cached hidden states rather than the model parameters.
We test HSO with pretrained Transformer-XL and GPT-2 language models, finding improvement on the WikiText-103 and PG-19 datasets in terms of perplexity, especially when evaluating a model outside of its training distribution. We also demonstrate downstream applicability by showing gains in the recently developed prompt-based few-shot evaluation setting, again with no extra parameters or training data.
\end{abstract}

\section{Introduction}
%
Finetuning a pretrained transformer language model (LM)~\citep{vaswani2017attention,radford2018improving,peters-etal-2018-deep,devlin-etal-2019-bert} is now the default method for attacking a task in modern NLP. Due to the high cost of pretraining, much research has been focused on how better to apply the pretrained models, rather than just improving pretraining itself. However, even finetuning can be too costly, especially for models such as the 175 billion parameter GPT-3~\citep{brown2020language}. As such, researchers have sought low cost alternatives, such as finetuning a small set of auxiliary parameters~\cite{pmlr-v97-houlsby19a}, or more recently leaving the LM weights fixed and passing a textual context designed to elicit the desired behavior via token prediction, such as in~\citet{brown2020language}.

One direction for language modeling in particular is to leave the LM parameters fixed, but update its intermediate quantities (e.g., \citealp{Dathathri2020Plug} and \citealp{qin-etal-2020-back}).
In this paper, we introduce Hidden-State Optimization (HSO), a method that contributes to this line of work. 
HSO 
first computes the language modeling loss as usual, then modifies the LM hidden states using the gradient of the loss (but critically reports the original loss). This process is repeated for each window of 10-25 tokens, updating the cached hidden states each time. Attending to these modified hidden states creates higher quality predictions for future tokens.

As an 
example of how future information can help embed past tokens, consider the garden path sentence: ``The old man the boat.'' The embedding for ``man'' will only depend on ``The'',  ``old'', and ``man'', so it will not 
reflect that ``man'' is being used as a verb. 
HSO can be seen as a method of incorporating future information into the representation of a context while still using a left-to-right LM.
BERT~\citep{devlin-etal-2019-bert} showed that bidirectional information passing improves embedding quality, which suggests that doing so should improve performance on downstream tasks.

We demonstrate HSO in the setting of language model evaluation on the WikiText-103 \citep{wt103} and PG-19~\citep{pg19} corpora, and find  
improvements in measured perplexity. In order to demonstrate that this translates into value for downstream applications we apply HSO 
to few-shot classification with the 1.5B parameter GPT-2, and find improvement in that setting as well.

\section{Related Work}
\textbf{Learning during inference.} 
HSO is related to methods that perform learning on the test set. 
One such method is dynamic evaluation (DE) \citep{krause2018dynamic,krause2019dynamic}, which was the inspiration for HSO. DE consists of using test inputs for learning after evaluating on them, which means a larger test set will result in a larger gain from its use.
This is not reflective of the small amount of text present in a setting such as conditional generation or few-shot classification, while using HSO for LM evaluation is. HSO is also cheaper than DE because it differentiates with respect to hidden states rather than the model parameters.
See Section~\ref{hso_vs_de} for more discussion and results on this point.
\vspace{-0.1cm}
\paragraph{Gradient-Based Optimization of Hidden States.}
\citet{qin-etal-2020-back} proposed Delorean, a method that incorporates future tokens into LM predictions by using backpropagation into earlier intermediate vectors. However, their goal is to produce better generations for intermediate timesteps, using sampled intermediate tokens and ground truth future tokens. We instead use the LM loss to tune past hidden states to allow better prediction of unseen future tokens. They also only perform gradient updates to logits while we update hidden states. 

Plug-and-Play language models (PPLM; \citealp{Dathathri2020Plug}) modify the behavior of pretrained LMs 
by updating hidden states at inference time, but with the goal of controllable generation (e.g., controlling sentiment) rather than improved fidelity. Unlike HSO, PPLMs require an attribute classifier which must be trained with labeled data. Several methods have been developed to more efficiently achieve the same goal as PPLM \citep{madotto-etal-2020-plug,krause-20}, and these ideas could potentially be applied in analogous ways to speed up HSO. 

\vspace{-0.1cm}
\paragraph{Alternatives to finetuning.} 
Our method is 
related to those that 
reduce the computational cost of finetuning by updating a smaller number of parameters or avoid finetuning altogether. 
\citet{pmlr-v97-houlsby19a} introduce adapter modules which are finetuned in lieu of the full model. 
\citet{DBLP:journals/corr/abs-2101-00190} introduce prefix-tuning, which adds a fixed set of learnable vectors to the beginning of the input sequence. The latter is related to 
%
using prompts for contextual generation, which has gained popularity both 
to extract 
information from 
language models (e.g., \citealp{radford2019language}, \citealp{jiang-etal-2020-know}) 
and perform tasks directly without updating any model parameters 
\citep{brown2020language}. Follow-up work has sought to understand the effectiveness of prompting \citep{le-scao-rush-2021-many} and automatically find or learn better prompts \citep{shin-etal-2020-autoprompt,DBLP:journals/corr/abs-2101-06804,qin-eisner-2021-learning}. 


\section{Method}
Let $f$ be a transformer language model computing the distribution for token $x_{t}$ given tokens $x_{1:t-1}$:
\vspace{-0.15cm}
\begin{equation*}
p_{t} = f(x_{1:t-1})
\end{equation*}
In practice, one may cache the hidden states, $\bm{h}_{t} \in \mathbb{R}^{\ell \times d}$, where $\ell$ is the number of layers and $d$ is the embedding size.
We represent this 
by factoring $f$ into
$f_h$ which computes hidden states (possibly depending on past hidden states) and $f_p$ which computes output probabilities from the hidden states:\vspace{-0.15cm} 
\begin{align}
\bm{h}_t &= f_h\left(x_t, \bm{h}_{1:t - 1}\right) \label{eq:hidden}\\
p_{t} &= f_p\left(\bm{h}_t\right)\nonumber
\end{align}
Given a loss function $L$ which takes as arguments the ground truth next word and a distribution over word types, one can then compute its gradient with respect to both the present hidden states $\bm{h}_t$, and with respect to the cached hidden states $\bm{h}_{1:t - 1}$:\vspace{-0.15cm}
\begin{align*}
\bm{g}_{\text{present}} &= \nabla_{\bm{h}_t} L\left(x_{t+1}, f_p(\bm{h}_{t})\right)\\
\bm{g}_{\text{cached}} &= \nabla_{\bm{h}_{1:t-1}} L\left(x_{t+1}, f_p(f_h(x_t, \bm{h}_{1:t - 1})\right)
\end{align*}
Denoting the concatenation of these two quantities along the time axis as $\bm{g}_t = \left[\bm{g}_\text{cached};\bm{g}_\text{present}\right]$, we can make a gradient update to the hidden states:\vspace{-0.15cm}
\begin{equation}\label{eq:update}
\tilde{\bm{h}}_{1:t} = \bm{h}_{1:t} - \eta\bm{g}_t
\end{equation}
where $\eta$ is the step size. We apply Adam \citep{kingma-15} to this update, but with modifications described in Section~\ref{adam}.

In practice, we use standard cross entropy as our loss function $L$. So, intuitively, we are updating the hidden states to make the actual word at position $t+1$ more likely under the language model's distribution $p_t$ by altering only the previously computed hidden states. Note that when we update the hidden states with gradient-based updates, it will no longer be the case that the set of hidden states follow the feedforward procedure defined by the architecture of the transformer language model. 

While computing the hidden state for $x_{t+1}$, we then substitute $\tilde{\bm{h}}_{1:t}$ into Eq.~\ref{eq:hidden} in place of  $\bm{h}_{1:t-1}$:\vspace{-0.15cm}
\begin{equation*}
    \bm{h}_{t+1} = f_h\left(x_{t+1}, \tilde{\bm{h}}_{1:t}\right)
\end{equation*}
Provided that the loss for timestep $t$ is computed with the unmodified hidden state $\bm{h}_t$ rather than $\tilde{\bm{h}}_t$, this may be done at test time 
without the loss being improved by ``looking into the future.''
We continue to update all hidden states at each step.\footnote{$\modstate{1}{t}$ is then a concatenation of hidden states which have been updated between 1 and $t$ times.}

In practice taking a gradient step after 
each token is too costly, so we can process blocks of $k$ tokens (which we will refer to as a \emph{window size} of $k$):\vspace{-0.2cm}
\begin{align*}
\bm{h}_{t+1} &= f_h\left(x_{t+1}, \modstate{1}{t}\right)\\
p_{t+1} &= f_p\left(\bm{h}_{t+1}\right)\\
\bm{h}_{t+2} &= f_h\left(x_{t+2}, [\bm{h}_{t+1:t+1}; \modstate{1}{t}]\right)\\\vspace{-1.8cm}
&\vdots \\
\bm{h}_{t + k} &= f_h\left(x_{t + k}, [\hstate{t+1}{t+k-1}; \modstate{1}{t}]\right)\\
p_{t + k} &= f_p(\bm{h}_{t+k})
\end{align*}
This sequence of computations is done in a single forward pass, but we have broken it up by token to make clear how a mix of unmodified and modified hidden states is used to embed each token in the window. Once the loss function, $L$, is applied to $x_{t+2:t+k+1}$ and $p_{t+1:t+k}$, a backwards pass is done to compute the gradient of the sum of the losses with respect to the hidden states, at which point the modified hidden states $\modstate{1}{t+k}$ are computed. 


$k$ has a twofold effect on computational cost, as it controls both the number of gradient steps and the number of tokens processed at a time. A very small $k$ will require many more forward passes and will not take advantage of GPU parallelism. 

\subsection{Modifications to Adam}\label{adam}
One way of applying Adam to the HSO gradient update would be to view the past hidden states as a single $T \times \ell \times d$ tensor, where $T$ is the maximum context size. This would allow use of just two moment estimate tensors $\bm{m}, \bm{v} \in \mathbb{R}^{T \times \ell \times d}$. This version of Adam performs very poorly, as a given value in the hidden state cache will not be consistently associated with the same moment estimate.

Instead, we keep first and second moment estimates $\bm{m}_i$ and $\bm{v}_i$ for each hidden state, discarding them once the corresponding hidden states are further in the past than the maximum attention length. This also requires maintaining a different optimizer step value for each block of $k$ hidden states, as Adam's bias correction depends on how many updates have been made to a moment estimate. In terms of implementation, we do not actually keep a separate vector for each hidden state, but pack them into a tensor which is translated along with the cached hidden state tensor.

\section{Experiments}
We demonstrate HSO with the Transformer-XL (TXL)~\citep{dai-etal-2019-transformer} and GPT-2\footnote{For GPT-2, we backpropagate into the key and value vectors rather than the full embeddings at each layer for ease of implementation. They differ by only a linear transformation, so we do not expect this to be a critical difference.} \citep{radford2019language} models implemented using FLAX~\citep{flax} and Haiku~\citep{haiku}, on top of JAX \citep{jax}. The TXL model is initialized from the HuggingFace Transformers~\citep{hftransformers} model trained on WikiText-103 (WT-103). The GPT-2 models are initialized from the OpenAI checkpoints.

\subsection{Language modeling}
We test HSO with the 
TXL and 345M parameter GPT-2 models on the pre-tokenized WikiText-103~\citep{wt103} and PG-19~\citep{pg19} datasets. As the TXL was trained on WT-103, this covers both an in-distribution and out-of-distribution (OOD) evaluation for it. We found that TXL was not stable in the OOD setting, but that  
resetting its hidden states to zeros upon reaching its maximum context size reduced the baseline perplexity significantly. We do not do this for HSO as it does not appear to need this stabilization. 
We evaluate GPT-2 with non-overlapping contexts for efficiency. The perplexities reported are per token, which differs between GPT-2 and the word based TXL. Out of vocabulary words are UNK-ed for TXL, but GPT-2 has an open vocabulary.

We used a window size of $k=25$, a learning rate of 0.003, and 0.65/0.9 for Adam's $\beta_1$ and $\beta_2$ parameters. We found that some HSO hyperparameter settings gave better performance, especially for GPT-2, but for the sake of parsimony report our main results with consistent hyperparameters.

\begin{table}[t]
\small
\centering
    \begin{tabular}{cll}
        Method & WT-103 & PG-19\\
        \midrule
        Baseline & 21.3/22.4 & 166.4/164.2\\
        HSO & \textbf{20.7}/\textbf{21.7} & \textbf{140.0}/\textbf{145.7}\
    \end{tabular}
    \caption{Language modeling validation/test perplexity with Transformer-XL (pretrained on WT-103). Importantly, PG-19 is out of distribution for this model.}
    \label{tab:transfoxl}
\end{table}

\begin{table}[t]
\small
\centering
    \begin{tabular}{cll}
        Method & WT-103 & PG-19\\
        \midrule
        Baseline & 21.5/20.7 & 26.7/\textbf{26.5} \\
        HSO & \textbf{21.0}/\textbf{20.3} & \textbf{25.1}/26.5\\
    \end{tabular}
    \caption{Language modeling validation/test perplexity with GPT-2 (345M parameters).}
\end{table}

Our LM results are shown in Tables~\ref{tab:transfoxl} and \ref{tab:gpt2}. HSO yields about a half a point improvement in perplexity on WT-103 with both architectures. While this is not a large improvement, recall that GPT-2's hidden states are reset every 1024 tokens, so this represents improvement in prediction within the context of one attention window, rather than cumulative training on the test set as in DE.

On PG-19, the perplexity improvements are larger for the most part: 1.6 points for GPT-2 on the validation set and over 10 points for TXL 
(but a <0.1 point increase for GPT-2 on the test set). As we used the same hyperparameters for all LM evaluations, HSO seems to be fairly robust to the choice of architecture and dataset. 

\subsubsection{Modifying HSO}\label{hparams}
Table~\ref{tab:hparams} shows the effect of various modifications to HSO on GPT-2's perplexity on the PG-19 validation set. Tuning Adam's parameters decreases perplexity by another point. Surprisingly, only updating the most recent window's hidden states (``present-only'') improves perplexity on PG-19 (initial experiments on WT-103 did not find this to be the case). This also requires significantly less computation. Since Adam tries to estimate moments over many steps this might seem to imply it is not necessary. To investigate this, we tested stochastic gradient descent (SGD) with several learning rates but it
performed worse than Adam for both full and ``present-only'' updates.\footnote{On the first step, Adam updates in the $L_\infty$ steepest descent direction so it differs from SGD even for only one step.}

\begin{table}[t]
\centering
\small
\begin{tabular}{lc}
Modifications & Perplexity\\
\midrule
None & 25.1\\
$\eta=3 \times 10^{-4}$, $\beta_1 = 0.8$ & 23.8\\
present-only & 23.6\\
$k=10$ & 24.4\\
$k=10$, present-only & 22.1\\
SGD, $\eta=0.01$, & 24.7\\
SGD, $\eta=0.01$, present-only & 25.1\\
\end{tabular}
\caption{GPT-2 (345M) perplexity on the PG-19 validation set. $\eta$ is learning rate, $k$ is window size, ``present-only'' means only the last $k$ hidden states are updated. 
}\label{tab:hparams}
\end{table}

\begin{table}[t]
\setlength{\tabcolsep}{5pt}
\centering
\small
\begin{tabular}{llcccc}
Dataset & $n$& \multicolumn{4}{c}{Method}\\
\cmidrule{3-6}
&& Baseline & DE\footnotemark & HSO & HSO-2\\
\midrule
\multirow{4}{*}{SST-2} & 2 & 53.9 & 52.2 (55.1) & 59.5 & \textbf{64.0}\\
& 4 & 58.3 & 55.6 (58.8) & 63.1 & \textbf{66.5}\\
& 6 & 57.9 & 56.2 (59.4) & 68.0 & \textbf{69.2}\\
& 8 & 58.4 & 59.9 (61.8) & \textbf{70.2} & 70.2\\
\midrule
\multirow{4}{*}{AGNews} & 2 & 53.1 & 32.2 (35.0) & 52.6 & \textbf{54.3}\\
& 4 & \textbf{77.8} & 52.2 (55.2) & 77.2 & 77.6\\
& 6 & 64.8 & \textemdash & 65.8 & \textbf{66.2}\\
& 8 & 63.3 & \textemdash & 68.5 & \textbf{69.3}\\
\end{tabular}
\caption{Effect of updating hidden states on few-shot classification accuracy of GPT-2-XL on SST-2 and AGNews, where $n$ is the number of examples per prompt. Neither hidden states or weights are updated for the baseline. HSO-2 is HSO with two gradient steps per window of text.}\label{tab:fewshot}
\end{table}
\footnotetext{Due to the much higher running time for using dynamic evaluation, these are partial results from running on a random subset of the test set. The accuracy in parentheses is a hypergeometric 95\% upper confidence bound. Future versions of this paper will have the full results. Furthermore, we exclude $n=6,8$ for AGNews due to running out of GPU memory on those input sizes.}

\subsection{Few-shot classification}
While 
HSO can give gains in perplexity, 
we would like to see whether it benefits other tasks as well. 
So, we consider few-shot learning from examples 
in the LM's 
context, as in GPT-3~\cite{brown2020language}.
Lacking GPT-3 access, we demonstrate our method with the 1.5B parameter GPT-2-XL model. 

We use the binary SST-2~\citep{socher2013recursive} and 4-way  AGNews~\citep{zhang2015character} classification datasets. 
We follow choices made by \citet{zhao2021calibrate}, including their prompt formats, but we made several changes to their procedure 
to reduce computational requirements and variance. 
Most importantly, we resampled a class-balanced prompt for every test example (but kept the prompt fixed between the baseline and HSO) rather than using a fixed prompt.\footnote{\citet{zhao2021calibrate} reported 
high variance 
based on prompt choice, 
so we made this choice in order to only need to run each evaluation once. The other two changes 
were to sample 1200 examples from the AGNews test set to expedite the evaluation, and to only use examples with  
$\leq$35 tokens in our prompts to reduce the required memory.} 
We used a learning rate of 0.01 and a window size of 10 tokens. Our experiments used a 24GB NVIDIA Quadro RTX 6000 GPU.

\davis{Can you take a look at the stuff I've added about DE here?} We also test DE, as in contrast to the LM setting, the amount of fine-tuning data will be the same between DE and HSO. We found that the learning rate of 0.01 led to the model collapsing to constant predictions, so we use a learning rate of $10^{-4}$ instead. We update the model every 10 tokens as with HSO, and recompute the hidden states after each update since the weights which produced them are no longer the model weights.

There are a few options to pick between when deciding what it meant to apply DE to this setting. One could choose to make a single gradient step based on the entire prompt, update the weights every 10 tokens but not recompute the hidden states, or perform multiple updates on the whole prompt. We chose what we believed was the closest comparison between HSO and DE, but did not experiment with these other variations. 

\subsubsection{Results}

Table~\ref{tab:fewshot} shows our results. HSO with a single gradient step leads to consistent improvements in accuracy across prompt sizes, and larger improvement with more prompt examples. The exceptions are AGNews with 2 and 4 example prompts, for which there is a slight decrease in accuracy. DE has similar performance to the baseline on SST-2, and degrades significantly on AGNews.

A longer prompt means both more examples to learn from and more gradient steps, so to disentangle the effect, we also tried two gradient steps per window (last column). 
This yields further improvement in 7 out of 8 cases. 
Surprisingly, for the cases where one gradient step was harmful, a second gradient step increases accuracy 
rather than causing further degradation.
%
Also, a second gradient step generally causes a larger increase in accuracy for shorter prompts (e.g., for SST-2, 
two steps with two examples 
beats one step with four examples). 

\subsubsection{Compute costs for HSO and DE}\label{hso_vs_de}
As we noted earlier, DE is not intended to be applied to a very small amount of text, so this is not an apples-to-apples comparison of methods, but can still help emphasize the differences between the two. In this setting, DE uses a much smaller amount of data (less than a single full GPT-2 window) to make updates to the entire transformer's weights. As such, it is not surprising it does not improve greatly over the baseline.

In terms of memory, the parameters and Adam moment estimates for DE of GPT-2-XL require more than 18GB in total. As the parameters are updated separately for each example, batching multiplies this overhead by the batch size, making DE infeasible for use on prompts coming from different distributions. HSO's extra overhead is the moment estimates for the hidden states, which cost \textasciitilde 1.2MB per token of input, for a total of \textasciitilde 1.3GB on a maximum size input. Furthermore, DE requires storing an additional copy of the model parameters, as they must be reset after each example. To avoid storing this extra copy on the GPU, we transferred it from RAM to GPU memory each time.

While the primary performance advantage over DE is reduced overhead and batching, we examine runtimes for each method in Table~\ref{tab:speed}. We additionally benchmark the 345M parameter GPT-2 for a speed comparison without the extra parameter transfer to the GPU. It is important to note that taking a single step per example instead of once per $k$ tokens would be much faster than either method, as both DE and HSO require $\lceil \frac{N}{k}\rceil$ backward passes for a length $N$ input.

\begin{table}[th]
\centering
\begin{tabular}{lccc}
     Method               & $n$ & \multicolumn{2}{c}{GPT-2 parameters}\\
     \cmidrule{3-4}
                          &     & 345M & 1558M\\
     \midrule
     \multirow{2}{*}{DE}  & 2   & 1.1  & 11.7\\
                          & 8   & 3.3  & 30.6\\
     \midrule
     \multirow{2}{*}{HSO} & 2   & 0.4  & 2.2\\ 
                          & 8   & 1.0  & 6.6\\
\end{tabular}
\caption{Seconds per example for few-shot evaluation using HSO and DE on SST-2. Because DE with GPT-2-XL requires copying the parameters from RAM to GPU memory every step, we also include speeds for GPT-2-medium which does not have that additional overhead.}\label{tab:speed}
\end{table}

\section{Conclusion and Future Work}
We presented a method that optimizes transformer language model hidden states, which improves LM perplexity and prompt-based few-shot classification, without additional parameters or data.

Future work will explore improving the cost of HSO by further investigation into updating only a subset of hidden weights, and approximation of the exact gradient update. Other directions we will explore are its application to conditional generation by improving the representation of the context, and its interaction with other methods for improving prompt-based few-shot classification.

\section*{Acknowledgements} Thank you to the reviewers for their time and feedback, which helped us to improve the paper. This material is based upon work supported by the National Science Foundation under Award No.~1941178.

\bibliography{anthology,custom}
\bibliographystyle{acl_natbib}
\end{document}